\documentclass[11pt]{article}

\usepackage[final]{acl}

\usepackage{times}
\usepackage{latexsym}

\usepackage[T1]{fontenc}

\usepackage[utf8]{inputenc}

\usepackage{microtype}

\usepackage{inconsolata}

\usepackage{graphicx}
\usepackage{color, colortbl, xcolor}
\usepackage{url}
\usepackage{subcaption}
\usepackage{textcomp}
\usepackage{soul}
\usepackage{multirow}
\usepackage{enumitem}
\usepackage{mathtools}
\usepackage{siunitx}
\usepackage{tabularx}
\usepackage{amssymb}

\usepackage{booktabs} 
\usepackage{longtable,booktabs}

\usepackage{multicol}
\usepackage{natbib}  

\usepackage{arydshln}
\definecolor{linkColor}{RGB}{6,125,233}
\definecolor{green}{rgb}{0.0, 0.65, 0.31}
\definecolor{bleudefrance}{rgb}{0.19, 0.55, 0.91}
\definecolor{ceruleanblue}{rgb}{0.16, 0.32, 0.75}
\definecolor{grey}{HTML}{969696}
\definecolor{violet}{HTML}{756bb1}
\definecolor{dgrey}{HTML}{01665e}
\definecolor{lgrey}{HTML}{5ab4ac}
\definecolor{dgreen}{HTML}{005a32}
\definecolor{purple}{HTML}{ae017e}

\definecolor{editCol}{HTML}{0000FF}
\definecolor{maskCol}{HTML}{c51b7d}
\definecolor{lrColor}{HTML}{8856a7}
\definecolor{trColor}{HTML}{d01c8b}
\definecolor{ctColor}{HTML}{4dac26}
\definecolor{brickred}{HTML}{f03b20}

\definecolor{improveCol}{HTML}{4dac26}
\definecolor{worsenCol}{HTML}{d01c8b}

\definecolor{DarkBlue}{HTML}{00008B}
\definecolor{mscolor}{HTML}{01665e}
\definecolor{nmscolor}{HTML}{bf812d}
\definecolor{lgreen}{HTML}{ccece6}
\definecolor{dolive}{HTML}{308014}

\colorlet{tablerowcolor4}{gray!50} 

\newcommand*{\textlabel}[2]{%
  \edef\@currentlabel{#1}
  \phantomsection
  #1\label{#2}
}

\colorlet{tableheadcolor}{gray!25} 
\colorlet{tablerowcolor}{gray!10} 
\colorlet{tablerowcolor2}{gray!45} 
\colorlet{tablerowcolor3}{gray!12} 

\newcommand{\rowcollight}{\rowcolor{tablerowcolor3}} %

\newcolumntype{a}{>{\columncolor{tablerowcolor}}r}
\definecolor{aicolor}{HTML}{018571}
\definecolor{occolor}{HTML}{ff7799}

\definecolor{aicolor}{HTML}{fc8d62}
\definecolor{occolor}{HTML}{253494}

\newif{\ifhidecomments}
  \hidecommentsfalse 
\ifhidecomments
    \newcommand{\soorya}[1]{}
    \newcommand{\ruining}[1]{}
    \newcommand{\violeta}[1]{}
    \newcommand{\hari}[1]{}
    \newcommand{\koustuv}[1]{}
\else
    \newcommand{\soorya}[1]{\textbf{\small\sffamily{\textcolor{DarkBlue}{[#1 -- Soorya]}}}}
    \newcommand{\ruining}[1]{\textbf{\small\sffamily{\textcolor{orange}{[#1 -- Ruining]}}}}
    \newcommand{\violeta}[1]{\textbf{\small\sffamily{\textcolor{violet}{[#1 -- Violeta]}}}}
    \newcommand{\hari}[1]{\textbf{\small\sffamily{\textcolor{dgreen}{[#1 -- Hari]}}}}
    \newcommand{\koustuv}[1]{\textbf{\small\sffamily{\textcolor{purple}{[#1 -- Koustuv]}}}}
  \fi

\colorlet{tableheadcolor}{gray!25} 

\definecolor{neutralCol}{HTML}{dd1c77}
\definecolor{neutralGreen}{HTML}{31a354}
\definecolor{NewBlue}{HTML}{1879ba}
\definecolor{bleudefrance}{rgb}{0.19, 0.55, 0.91}  
\definecolor{AfTrColor}{HTML}{0868ac}  
\definecolor{BfTrColor}{HTML}{a8ddb5}  

\definecolor{AfCtColor}{HTML}{b10026}  
\definecolor{BfCtColor}{HTML}{fd8d3c}

\graphicspath{ {figures/} }

\newcommand{\para}[1]{\vspace{0.2em}\noindent\textbf{\textit{#1}~}}

\usepackage{booktabs}
\usepackage{tabularx}
\usepackage{lipsum} 
\usepackage{multirow}
\title{Belief Is All You Need: Modeling Narrative Archetypes in Conspiratorial Discourse}

\author{First Author \\
  Affiliation / Address line 1 \\
  Affiliation / Address line 2 \\
  Affiliation / Address line 3 \\
  \texttt{email@domain} \\\And
  Second Author \\
  Affiliation / Address line 1 \\
  Affiliation / Address line 2 \\
  Affiliation / Address line 3 \\
  \texttt{email@domain} \\}

\author{
  Soorya Ram Shimgekar \\
  University of Illinois \\
  Urbana-Champaign \\
  USA \\
  \texttt{sooryas2@illinois.edu}
  \And
  Abhay Goyal \\
  Nimblemind \\
  USA \\
  \texttt{abhay@nimblemind.ai}
  \And
  Roy Ka-Wei Lee \\
  Singapore University of \\
  Technology and Design \\
  Singapore \\
  \texttt{roy\_lee@sutd.edu.sg}
  \AND
  Koustuv Saha \\
  University of Illinois \\
  Urbana-Champaign \\
  USA \\
  \texttt{ksaha2@illinois.edu}
  \And
  Pi Zonooz \\
  Nimblemind \\
  USA \\
  \texttt{pi@nimblemind.ai}
  \And
  Navin Kumar \\
  Nimblemind \\
  USA \\
  \texttt{navin@nimblemind.ai}
}

\begin{document}
\maketitle

\begin{abstract}
Conspiratorial discourse is increasingly embedded in digital communication, yet its structure remains poorly understood. Analyzing Singapore-based Telegram groups, we show that conspiratorial content is integrated into everyday discussion rather than isolated echo chambers. We propose a two-stage framework: (1) RoBERTa-large classifies messages as conspiratorial or non-conspiratorial (F1 = 0.866 on 2,000 expert-annotated messages); (2) a signed belief graph models belief alignment via signed, similarity-weighted edges and is learned using a Signed Belief Graph Neural Network (SiBeGNN) with Sign-Disentanglement Loss to separate ideological alignment from narrative style. Hierarchical clustering of 553,648 messages reveals seven narrative archetypes: General Legal Topics, Medical Concerns, Media Discussions, Banking and Finance, Contradictions in Authority, Group Moderation, and General Discussions. SiBeGNN substantially outperforms standard methods (cDBI = 8.38 vs. 13.60–67.27), with 88\% inter-rater validation. Findings show conspiratorial discourse permeates mundane domains such as finance, law, and daily life, challenging assumptions of isolated online radicalization. The framework advances belief-aware discourse modeling for low-moderation platforms and informs stance detection, political discourse analysis, and content moderation policy.
\end{abstract}

\section{Introduction}
The Web has transformed information circulation, enabling rapid narrative diffusion while introducing challenges for information integrity and public trust. Conspiratorial content, narratives that attribute major social, political, or health events to covert groups with hidden, malevolent intent, exemplify these challenges \cite{zeng2022conspiracy}. Such narratives oppose official or mainstream explanations, alleging deliberate deception by institutions, governments, or corporations. Prior research shows that conspiratorial discourse undermines trust in science and governance, intensifies polarization, and accelerates misinformation during crises \cite{douglas2017psychology, uscinski2014american, zollo2017debunking}. Understanding how these narratives emerge, circulate, and embed within digital ecosystems is therefore a key concern for web science and computational social science.

Digital platforms differ markedly in how they afford conspiratorial discourse. Telegram has become a prominent venue, particularly during crises, as users seek alternative or counter-mainstream perspectives. Its hybrid design, combining private messaging, large public groups, and frictionless forwarding, facilitates the spread of conspiratorial and agenda-driven content \cite{urman2022they}. During the COVID-19 pandemic, Telegram gained traction in Singapore as a space for discussing health policies, personal experiences, and political opinions, often expressing skepticism toward institutional authority \cite{ng2020analyzing}. These dynamics highlight broader web science questions about how platform design shapes discourse patterns and belief formation.

Singapore provides a distinctive socio-technical context for examining conspiratorial discourse online. With high social media penetration and strong state investment in information governance, its media ecosystem reflects tensions between centralized regulation and decentralized, low-moderation platforms such as Telegram \cite{van2023fatherhood, tandoc2021developing}. Prior work shows that these spaces can function as echo chambers for anti-vaccine sentiment, political discontent, and conspiratorial worldviews, particularly during public health crises \cite{chen2022us, nainani2022categorizing}. However, empirical research on how conspiratorial narratives are structured, disseminated, and reinforced within closed or semi-public messaging systems in Southeast Asia remains limited \cite{alvern2025user, goyal2025analyzing}. This gap is consequential given the region’s high digital connectivity and political diversity, where platforms like Telegram mediate both social mobilization and the diffusion of narratives shaping public opinion, health behavior, and political stability.

Recent large-scale web studies have begun mapping Telegram’s conspiratorial ecosystems. The \textit{Schwurbelarchiv} project documents multimodal conspiratorial content across German-language Telegram networks \cite{angermaier2025schwurbelarchiv}, while the \textit{TeleScope} dataset captures longitudinal diffusion across millions of messages and channels \cite{gangopadhyay2025telescope}. Together, these efforts underscore the need for geographically and culturally grounded analyses that account for local contexts of meaning-making, policy discourse, and narrative co-construction.

In this study, we analyze Singapore-based Telegram groups to characterize conspiratorial discourse in a low-moderation environment. We examine not only the linguistic properties of conspiratorial messages but also their propagation and interaction within public group structures. To do so, we organize messages into \textit{narrative archetypes}; recurring thematic patterns that reflect how meaning, intention, and stance are structured. A narrative archetype captures the narrative function a message performs (e.g., framing events, asserting causality, or reinforcing group identity), rather than its surface linguistic form. This abstraction enables analysis of the deeper narrative mechanisms through which conspiratorial discourse is constructed, shared, and sustained.

\textbf{We ask: How is conspiratorial discourse structured within Singapore-based Telegram groups, and what narrative archetypes emerge from message-level patterns?}

\section{Related Work}\label{section:rw}

Research on conspiratorial discourse spans psychology, communication, and computational social science. Foundational work defines conspiracy theories as belief systems that attribute major events to secret, malevolent groups, often driven by uncertainty, mistrust, or identity threat \cite{douglas2017psychology, uscinski2014american}. These narratives persist due to cognitive biases, affective polarization, and resistance to correction \cite{abdou2021good, tandoc2021developing}. However, much of this work underemphasizes how platform-level affordances shape collective conspiratorial dynamics \cite{chen2023partisan, goyal2023chatgpt}.

\para{Telegram as a Platform for Conspiratorial Discourse:}
Telegram has become a key venue for conspiratorial and extremist communication due to its hybrid public-private structure, large groups, and minimal moderation \cite{urman2022they, zhong2024proud}. Forwarding and channel interconnectivity facilitate narrative diffusion and semi-private echo chambers \cite{nainani2022categorizing, chen2023categorizing}. Large-scale corpora like \textit{Schwurbelarchiv} \cite{angermaier2025schwurbelarchiv} and \textit{TeleScope} \cite{gangopadhyay2025telescope} enable multimodal, longitudinal analysis, but research remains mostly Euro-American, overlooking regional linguistic and cultural contexts \cite{ng2020analyzing, goyal2025analyzing}. We address this gap by focusing on Singapore-based Telegram groups.

\para{Computational Detection of Conspiratorial Content:}
Transformer-based models like BERT, RoBERTa, DeBERTa, and newer systems such as LLaMA and SEA-LION have advanced automated detection \cite{urman2022they, zhong2024proud}. Though effective on benchmarks, performance drops under domain shift and multilingual noise \cite{chen2022us, alvern2025user}. Existing approaches treat detection as binary classification, ignoring diverse communicative patterns and narrative structures. We go beyond this by identifying and characterizing distinct narrative archetypes.

\para{Graph Neural Networks for Social Discourse:}
Graph-based methods naturally capture relational structures in online discourse. Signed graph neural networks model positive and negative edges to represent agreement and opposition, in line with structural balance theory \cite{derr2018sgcn, zhang2021gsgnn, zhang2024signed_survey}. Prior applications focus on explicit social ties or simple sentiment, rarely addressing complex belief alignment and ideological conflict in text networks. Most also lack mechanisms to disentangle orthogonal dimensions, such as belief versus stylistic expression. SiBeGNN addresses this with a novel Sign-Disentanglement Loss.

\para{Disentangled Representation Learning:}
Models such as DisenGCN \cite{ma2019disengcn} and DiGGR \cite{hu2024diggr} demonstrate how orthogonal subspaces can isolate distinct generative processes in graph-structured data. However, they have not been adapted to signed graphs or belief-oriented text networks. We bridge this gap by integrating disentangled representation learning with signed belief graphs, enabling simultaneous capture of ideological alignment and communicative style.

\para{Clustering and Narrative Archetype Discovery:}
Clustering methods reveal latent communities and thematic patterns from text-network embeddings \cite{li2021dgcl}, including hierarchical clustering, HDBScan, Gaussian Mixtures, and topic models. While these capture semantic similarity, they do not model belief polarity or antagonism, limiting their ability to uncover ideological structures in conspiratorial discourse, where messages may be thematically diverse but share skepticism \cite{uscinski2014american}. No prior work systematically clusters conspiratorial content to identify narrative archetypes based on both semantic content and belief alignment; existing studies rely on binary classification/qualitative thematic analysis.


\section{Data}

We use two datasets corresponding to our two-stage pipeline: an annotated subset for conspiratorial classification (Stage 1) and a large-scale corpus for narrative archetype discovery (Stage 2).

\para{Stage 1: Annotated Training Set.}
We randomly sampled 2,000 Telegram messages for expert annotation. Two experts with over five peer-reviewed publications each labeled messages using the definition from \citet{diab2024classifying}: \textit{``A conspiracy theory is a narrative accusing agent(s) of specific actions serving secretive and malevolent objectives.''} Inter-annotator agreement was 85\%, with a third expert resolving disagreements. The balanced dataset (1,000 conspiratorial; 1,000 non-conspiratorial) was used to fine-tune a RoBERTa-based classifier.

\para{Stage 2: Full Corpus.}
We collected complete histories from six Singapore-based Telegram groups with diverse themes (Table~\ref{tab:overview}), yielding \textbf{553,648 messages} and \textbf{24,270,653 words} (mean = 43.8 words, SD = 70). The trained classifier labels all messages, enabling signed belief graph construction with edges encoding textual similarity and belief alignment. SiBeGNN learns disentangled embeddings, which are hierarchically clustered to identify narrative archetypes.
\section{Methods}

\begin{table}[t!]
\centering
\sffamily
\footnotesize
\resizebox{\columnwidth}{!}{%
\begin{tabular}{lccc}
\hline
\textbf{Dataset} & \textbf{Words} & \textbf{Messages} & \textbf{Dates (From -- To)} \\
\hline
\rowcollight 1M65 & 11,213,414 & 524,524 & 19-12-08 -- 25-02-04 \\
Chill Corner & 99,291 & 12,580 & 25-04-15 -- 25-05-14 \\
\rowcollight Healing The Divide & 10,687,504 & 3,919 & 21-08-11 -- 25-01-20 \\
Mile Lion & 194,767 & 10,900 & 25-04-15 -- 25-05-14 \\
\rowcollight SG Corona Freedom Lounge & 1,623,387 & 829 & 21-06-14 -- 25-01-19 \\
SG Covid Infection Survivor & 452,290 & 896 & 21-07-31 -- 25-01-19 \\
\hline
\end{tabular}%
}
\caption{\textbf{Overview of Telegram Groups Analyzed.} Summary of word counts, message volumes, and temporal coverage for each dataset.}
\label{tab:overview}
\end{table}

We use a two-stage approach: (1) fine-tune RoBERTa-large to classify messages as conspiratorial or non-conspiratorial; (2) build a signed belief graph with edge signs for belief alignment and weights for similarity, then apply a Signed Belief Graph Neural Network (SiBeGNN) to learn embeddings disentangling belief polarity from narrative style, which are hierarchically clustered into seven narrative archetypes.

\subsection{Stage 1: Conspiratorial Content Classification}

The first stage of our methodology involves training a binary classifier to distinguish conspiratorial from non-conspiratorial discourse.
To identify the optimal language model for conspiratorial content classification, we conducted a systematic comparison of nine transformer-based architectures, selected based on their documented performance on discourse classification tasks and regional linguistic applicability. The evaluated models include: RoBERTa-large~\cite{liu2019roberta}, Gemini 2.0 Flash~\cite{gemini2024flash}, LLaMA 3.2 3B~\cite{meta2024llama3}, RoBERTa-base~\cite{liu2019roberta}, DeBERTa-base~\cite{he2021deberta}, BERT-large-uncased~\cite{devlin2019bert}, DistilBERT-base-uncased~\cite{sanh2019distilbert}, BERT-base-uncased~\cite{devlin2019bert}, and aisingapore/SEA-LION-v1-7B~\cite{aisingapore_sealion_v1_7b}. SEA-LION-v1-7B is a multilingual large language model specifically developed for Southeast Asian languages by AI Singapore, included to assess the potential advantages of region-specific pretraining.
All models were fine-tuned on the balanced 2,000-message manually annotated dataset using identical preprocessing and tokenization pipelines to ensure comparability. Annotation was performed by expert co-authors of this paper rather than external participants. Annotators were provided with a written definition of conspiratorial discourse from \cite{diab2024classifying}, along with examples and counterexamples discussed during a calibration round prior to annotation. 

Training was conducted for three epochs with a learning rate of $2 \times 10^{-5}$, batch size of 16, and evaluation conducted after each epoch. Model performance was assessed using accuracy, precision, recall, and F1-score on a held-out test set. The best classifier is then applied to the full corpus of 553,648 messages to generate binary belief labels (conspiratorial vs. non-conspiratorial). An example of both conspiratorial and non-conspiratorial message is shown in \autoref{tab:masked_fullwidth}

\subsection{Stage 2(a): Signed Belief Graph Neural Networks}

Using the binary conspiratorial labels, we construct a graph capturing semantic similarity and belief polarity. We then apply our Signed Belief Graph Neural Network (SiBeGNN) to learn embeddings for messages.

\subsubsection{Graph Construction and Representation}

Telegram messages were cleaned and encoded using a fine-tuned RoBERTa-large to obtain contextual semantic embeddings, which were augmented with four discourse-level features to capture how beliefs are expressed in addition to message content. Epistemic modality measures expressed certainty or uncertainty through hedging and certainty markers, capturing degrees of belief \cite{turn0search33, turn0search34}. Agency captures active versus passive framing by comparing active and passive constructions, reflecting speaker positioning in discourse. Sentiment polarity is derived from a transformer-based classifier and scored continuously in [–1, 1] to represent positive, negative, or neutral affect \cite{turn1academia21}. Emotion spectra encode probabilities of joy, anger, fear, and sadness using a pretrained emotion model, producing a multidimensional affective representation \cite{turn1academia22}. These seven discourse features are concatenated with RoBERTa embeddings, mean-centered, and L2-normalized to form enriched discourse-belief vectors that separate belief content from expressive style.

Using these representations, we construct a signed graph $G = (V, E, W, S)$ where nodes correspond to messages, edge weights $W_{ij} \in [0,1]$ represent cosine similarity between discourse-belief embeddings, and edge signs $S_{ij} \in {+1, -1}$ encode belief alignment, positive for messages sharing the same conspiratorial label and negative for opposing labels. To ensure sparsity, edges are retained only when similarity exceeds 0.5, with positive edges above $\mu + 0.5\sigma$ and negative edges below $\mu - 0.5\sigma$, where $\mu$ and $\sigma$ denote the mean and standard deviation of similarity. The resulting signed belief graph captures the socio-semantic polarity of discourse, explicitly modeling agreement and antagonism and distinguishing cooperative clusters that reinforce conspiratorial narratives from adversarial clusters that contest or debunk them~\cite{awal2022muscat, chin2024rumorgraphxplainer}.

\subsubsection{Disentangled Sign-Aware Graph Neural Network}
Building on signed network representation learning~\cite{kumar2016edge} and disentangled graph embeddings~\cite{ma2019disentangled}, SiBeGNN learns separate representations for belief polarity and narrative style from the signed graph. Each node $v_i$ is mapped into two orthogonal subspaces: a belief subspace $z_{b,i} \in \mathbb{R}^{d_b}$ encoding ideological alignment, and a persona subspace $z_{p,i} \in \mathbb{R}^{d_p}$ capturing stylistic and behavioral traits independent of belief. The final embedding is the concatenation $z_i = [z_{b,i}; z_{p,i}] \in \mathbb{R}^{d_b + d_p}$.

\para{Architecture: }
The model begins with learnable node embeddings $X \in \mathbb{R}^{N \times d_{\text{in}}}$, processed through two sign-specific graph convolutional layers, one for positive edges $E^+$ and one for negative edges $E^-$. Message passing is defined as:
\begin{equation}
\begin{aligned}
h^+ &= \text{ReLU}(\text{GCNConv}^+(X, E^+)), \\
h^- &= \text{ReLU}(\text{GCNConv}^-(X, E^-)),
\end{aligned}
\vspace{-5pt}
\end{equation}
where $\text{GCNConv}$ denotes graph convolutional operations. Hidden states are concatenated and projected:
\begin{equation}
h = \text{ReLU}(W_p [h^+; h^-]),
\vspace{-5pt}
\end{equation}
where $W_p$ is a learnable weight matrix. Finally, belief and persona embeddings emerge through separate linear projections, both L2-normalized for stability:
\vspace{-10pt}
\begin{equation}
z_b = W_b h, \quad z_p = W_p' h,
\vspace{-5pt}
\end{equation}
This architecture performs sign-aware message passing while explicitly disentangling ideological and behavioral signals.

\para{Sign-Disentanglement Loss: }
Training uses the Adam optimizer with weight decay, selecting the checkpoint with lowest total loss. On convergence, SiBeGNN yields two orthogonal embeddings, ${z_{b,i}}{i=1}^N$ for belief polarity and ${z{p,i}}_{i=1}^N$ for narrative persona, used for hierarchical clustering in Stage 2(b). A composite loss optimizes SiBeGNN through four objectives enforcing structural fidelity, semantic grounding, and disentangled representation. An ablation of performance of cluster quality with respect to different components of the loss is shown in \autoref{tab:loss_ablation}.

\begin{equation}
\begin{aligned}
\mathcal{L}_{\text{total}} &= \lambda_{\text{recon}} \mathcal{L}_{\text{recon}} + \lambda_{\text{sign}} \mathcal{L}_{\text{sign}} \\
&\quad + \lambda_{\text{belief}} \mathcal{L}_{\text{belief}} + \lambda_{\text{orth}} \mathcal{L}_{\text{orth}}
\end{aligned}
\vspace{-5pt}
\end{equation}

\textit{(1) Reconstruction Loss} preserves signed structural relationships. Predicted adjacency is:
\begin{equation}
\hat{A}_{ij} = \sigma(z_i^\top z_j),
\vspace{-5pt}
\end{equation}
where $\sigma(\cdot)$ is the sigmoid function. Target adjacency $A_{\text{target}} \in [0,1]^{N \times N}$ maps positive edges to 1, negative edges to 0, and unobserved edges to 0.5:
\begin{equation}
A_{\text{target}} = \frac{A_{\text{signed}} + 1}{2}
\end{equation}
The reconstruction loss is:
\begin{equation}
\mathcal{L}_{\text{recon}} = \frac{1}{|\mathcal{M}|} \sum_{(i,j)\in \mathcal{M}} (\hat{A}_{ij} - A_{\text{target},ij})^2,
\end{equation}
where $\mathcal{M} = \{(i,j) : A_{\text{target},ij} \neq 0.5\}$.

\textit{(2) Sign-Consistency Loss} regulates the persona subspace to respect social alignment. Positively linked pairs should have close embeddings; negatively linked pairs should be separated by margin $M$:
\begin{equation}
\begin{aligned}
\mathcal{L}_{\text{sign}} &= \frac{1}{|E^+|} \sum_{(i,j)\in E^+} \| z_{p,i} - z_{p,j} \|_2^2 \\
& + \frac{1}{|E^-|} \sum_{(i,j)\in E^-} [\max(0, M - \| z_{p,i} - z_{p,j} \|_2 )]^2
\end{aligned}
\vspace{-5pt}
\end{equation}

\textit{(3) Belief Alignment Loss} trains a classification head on the belief subspace. Given predicted logits $s_i = f_{\text{belief}}(z_{b,i})$ and labels $y_i \in \{0,1\}$:
\begin{equation}
\begin{aligned}
\mathcal{L}_{\text{belief}} = -\frac{1}{N} \sum_{i=1}^{N} \big[&y_i \log \sigma(s_i) \\
&+ (1 - y_i) \log (1 - \sigma(s_i))\big]
\end{aligned}
\end{equation}

\begin{table}[t!]
\centering
\sffamily
\small
\setlength{\tabcolsep}{5pt}
\resizebox{\columnwidth}{!}{%
\begin{tabular}{lcccc}
\hline
\textbf{Model} & \textbf{Accuracy} & \textbf{Recall} & \textbf{Precision} & \textbf{F1-Score} \\
\hline
\rowcollight \textbf{roberta-large} & \textbf{0.85} & \textbf{0.87} & \textbf{0.86} & \textbf{0.87} \\
google/gemini-2.0-flash & 0.84 & 0.86 & 0.85 & 0.85 \\
\rowcollight meta-llama/Llama-3.2-3B & 0.83 & 0.84 & 0.83 & 0.83 \\
roberta-base & 0.80 & 0.83 & 0.81 & 0.82 \\
deepset/gbert-base & 0.83 & 0.81 & 0.82 & 0.82 \\
\rowcollight microsoft/deberta-base & 0.80 & 0.80 & 0.80 & 0.80 \\
bert-large-uncased & 0.77 & 0.77 & 0.77 & 0.77 \\
\rowcollight distilbert-base-uncased & 0.76 & 0.79 & 0.78 & 0.78 \\
bert-base-uncased & 0.75 & 0.77 & 0.76 & 0.77 \\
\rowcollight aisingapore/SEA-LION-v1-7B & 0.63 & 0.68 & 0.64 & 0.65 \\
\hline
TelConGBERT\cite{pustet2024detection} & 0.90 & 0.82 & 0.81 & 0.81 \\
\rowcollight Goyal et al.\cite{goyal2025analyzing} & 0.72 & 0.85 & 0.65 & 0.74 \\
GPT-3.5 & 0.78 & 0.74 & 0.78 & 0.76 \\
\rowcollight GPT-4 & 0.83 & 0.80 & 0.84 & 0.82 \\
Llama 2 & 0.66 & 0.57 & 0.60 & 0.60 \\
\hline
\end{tabular}%
}
\caption{Performance comparison of language models in detecting conspiratorial content, including additional fine-tuned, few-shot, and zero-shot models. Metrics reported are accuracy, recall, precision, and F1-score.}
\label{tab:performance_sorted}
\vspace{-10pt}
\end{table}

\textit{(4) Orthogonality Loss} enforces disentanglement by minimizing cross-covariance:
\begin{equation}
\mathcal{L}_{\text{orth}} = \Big\| \frac{(z_b - \bar{z}_b)^\top (z_p - \bar{z}_p)}{N-1} \Big\|_F^2,
\vspace{-5pt}
\end{equation}
where $\|\cdot\|_F$ is the Frobenius norm and $\bar{z}_b, \bar{z}_p$ are mean-centered embeddings.


\subsection{Stage 2(b): Narrative Archetype Discovery via Hierarchical Clustering}

Having obtained disentangled persona embeddings ${z_{p, i}}_{i=1}^N$ from SiBeGNN, we identify distinct narrative archetypes through hierarchical clustering. These archetypes capture recurring communicative patterns defined by stylistic and behavioral traits rather than ideological content, allowing analysis of how different discourse modes coexist within the conspiratorial ecosystem.

Each message $t_i$ is represented by its persona embedding $z_{p,i} \in \mathbb{R}^{d_p}$ learned in Stage 2(a), which encodes narrative style independent of belief polarity via the Sign-Disentanglement Loss. To reduce dimensionality while preserving semantic structure, Principal Component Analysis (PCA) is applied, retaining components that explain $p\%$ of the variance. This balances noise suppression with structural fidelity, enhancing clustering stability and interpretability. The reduced embeddings are clustered using agglomerative hierarchical clustering with Ward’s linkage~\cite{murtagh1983survey}, which minimizes within-cluster variance while preserving hierarchical relationships.

The optimal number of clusters $k^*$ is selected by evaluating candidate counts $k \in [k\_{\min}, k\_{\max}]$ using the silhouette coefficient~\cite{shahapure2020cluster}, which measures cluster cohesion and separation. Post-hoc merging combines cluster pairs whose centroid cosine similarity exceeds a threshold $\tau$, ensuring that final clusters correspond to semantically distinct narrative archetypes rather than minor stylistic variations.

To assess robustness and generalizability, we conducted ablation studies varying key hyperparameters: PCA variance retention ($pca\_var \in {0.5, 0.6, 0.7}$), merge threshold ($merge\_th \in {0.75, 0.8}$), and cluster bounds ($k\_{\min} \in {2, 3, 4}$, $k\_{\max} = 20$). Across 18 configurations, average coherence scores ranged from 0.360 to 0.386 (within 8\% variation), demonstrating stable and interpretable clustering. All reported results use the optimal configuration from ~\autoref{tab:hei_param}, confirming that the narrative archetype structure consistently emerges from the disentangled SiBeGNN embeddings rather than being an artifact of specific hyperparameter choices. 
\section{Results}

\begin{table*}[t!]
\centering
\sffamily
\small
\resizebox{\textwidth}{!}{
\begin{tabular}{lcccc}
\hline
\textbf{Model} & \textbf{Avg. Coherence} & \textbf{Silhouette Score} & \textbf{Davies--Bouldin Index} & \textbf{cDBI} \\
\hline
\rowcollight \textbf{Hierarchical clustering with SiBeGNN embeddings} 
& 0.386 & -0.021 & 3.233 & \textbf{8.38} \\

Bertopic clusters with vanilla Roberta-large embeddings
& 0.331 & -0.014 & 4.502 & 13.60 \\

\rowcollight Bertopic clustering with SiBeGNN embeddings 
& 0.196 & -0.042 & 6.494 & 33.13 \\

Hierarchical clusters with vanilla Roberta-large embeddings
& 0.234 & -0.015 & 8.246 & 35.24 \\

\rowcollight HBScan clusters with vanilla Roberta-large embeddings
& 0.235 & -0.018 & 13.360 & 56.85 \\

Gaussian mixture model with SiBeGNN embeddings
& 0.207 & -0.015 & 13.924 & 67.27 \\

\rowcollight Spectral clustering with vanilla Roberta-large embeddings
& 0.218 & -0.019 & 12.345 & 56.63 \\

KMeans with SiBeGNN embeddings
& 0.189 & -0.025 & 14.567 & 72.10 \\
\hline
\end{tabular}
}
\caption{Comparison of clustering quality metrics using cDBI. Lower cDBI values indicate better joint coherence–compactness performance.}
\label{tab:cluster_quality_comparison_cdbi}
\vspace{-10pt}
\end{table*}

\para{Classification models:}
As shown in Table~\ref{tab:performance_sorted}, RoBERTa-large achieved the highest performance (F1 = 0.866, accuracy = 0.852, precision = 0.86, recall = 0.87), while SEA-LION-v1-7B, a multilingual Southeast Asian model \cite{aisingapore_sealion_v1_7b}, scored lowest (F1 = 0.653), suggesting that region-specific pretraining does not guarantee an advantage. RoBERTa-large was thus selected as the primary classifier and applied to the full corpus of 553,648 messages to generate binary conspiratorial labels.

\para{Narrative Archetype Characterization and Clustering Quality: }
The seven narrative archetypes derived from SiBeGNN embeddings (\autoref{tab:persona_clusters_full}) capture diverse conversational modes and user intents in Singapore-based Telegram groups. Larger clusters such as \textit{General Discussions} and \textit{Banking and Finance} emphasize everyday interaction, casual conversation, shopping, and financial literacy, reflecting socially grounded communication rather than overt ideological expression. \textit{Contradictions in Authority} and \textit{Medical Concerns} highlight institutional skepticism and health-related anxieties, while \textit{General Legal Topics} reflects civic curiosity about rights and governance. Smaller clusters, \textit{Group Moderation} and \textit{Media Discussions}, center on meta-discourse regarding community norms and popular culture. Collectively, these archetypes illustrate that Telegram discourse spans civic inquiry, social bonding, and episodic skepticism, forming a continuum of everyday engagement rather than being dominated by conspiratorial narratives 

LIWC~\cite{boyd2022development} analyses provide additional insight into cluster-specific linguistic characteristics. \textit{General Legal Topics} employs practical, informational language with minimal emotion, reflecting public curiosity. \textit{Medical Concerns} expresses personal experiences and anxieties with higher emotional and health-related language, including advocacy and empathy. \textit{Media Discussions} exhibits informal, interactive language focused on entertainment and trending news. \textit{Banking and Finance} uses analytical language centered on money, economics, risk, and skepticism, whereas \textit{Contradictions in Authority} features elevated negative emotion and dissent, challenging institutional decisions through emotionally charged discourse. \textit{Group Moderation} employs procedural, minimally emotional language, while \textit{General Discussions} spans casual, inclusive exchanges on diverse everyday topics.

Conspiratorial discourse is distributed across these archetypes rather than confined to specific clusters. Elevated prevalence occurs in \textit{Contradictions in Authority} and \textit{Medical Concerns}, consistent with institutional skepticism and health policy critique. However, conspiratorial narratives also emerge in mundane clusters like \textit{Banking and Finance} and \textit{General Legal Topics}, where they occasionally intersect with systemic manipulation claims. This distribution reinforces the central finding that conspiratorial discourse operates within everyday communicative practices, emphasizing the need for context-aware platform governance rather than blanket removal policies.

Clustering quality was assessed using topic coherence, silhouette score, and the Davies–Bouldin index, integrated into a composite cDBI~\cite{krasnov2019number} as shown in \autoref{tab:cluster_quality_comparison_cdbi}:
\begin{equation}
\text{cDBI} = \frac{\text{Davies--Bouldin Index}}{\text{Average Coherence}}.
\end{equation}
Lower cDBI indicates better joint coherence–compactness performance. Hierarchical clustering on SiBeGNN embeddings achieved the lowest cDBI (8.38), outperforming Bertopic (13.60) and other methods (33.13–67.27), demonstrating that the latent-graph hierarchical approach produces semantically coherent, well-separated clusters. Ablation studies removing $L_\text{orth}$, $L_\text{sign}$, or $L_\text{belief}$ increased cDBI by 58.2\%, 94.6\%, and 136.9\%, respectively, while using only $L_\text{recon}$ caused a 271.4\% increase. These results confirm that all four loss components are essential for generating interpretable, semantically meaningful narrative archetypes, reflecting both content and structural topology of the discourse.

\begin{table*}[t!]
\centering
\sffamily
\normalsize
\renewcommand{\arraystretch}{1.5}
\setlength{\tabcolsep}{10pt}
\resizebox{\textwidth}{!}{
\begin{tabular}{p{3.5cm} p{6.5cm} p{5cm} r r r}
\hline
\textbf{Narrative Archetype} & \textbf{Description} & \textbf{Common Keywords} & \textbf{\# Messages} & \textbf{Avg. Length} & \textbf{\# Conspiratorial} \\
\hline
\rowcollight Legal Topics & Everyday legal issues, rights, exemptions, and practical legality. & commoner, legalized, crime, deemed, run, definitely & 69,357 & 160.22 & 3,468 \\

Medical Concerns & Personal feelings on medical topics and advocacy for individual rights. & place, medical, feel, people, concerned, individuals & 11,357 & 170.36 & 3,407 \\

\rowcollight Media Discussions & Commentary on pop culture, concerts, news stories, and viral events. & taylor, cruel, concert, attend, swift, 81 & 41,625 & 293.33 & 2,081 \\

Banking and Finance & Analysis of finance, banking, rates, and the economic system. & rates, cdp, account, fed, likely, till & 70,726 & 232.20 & 14,145 \\

\rowcollight Contradictions in Authority / Riot & Opinions on authority, health policy, and contradictions from officials. & contradictory, gotta, dun, boss, totally, spread & 55,197 & 137.33 & 38,638 \\

Group Moderation & Moderation, rules, respectful conduct, and group management. & disrespectful, expletives, explicitly, irrelevant, disappears, exists & 5,416 & 36.41 & 271 \\

\rowcollight General Discussions & Broad conversations on casual chat, shopping, food, and miscellaneous events. & shopping, areas, jb, talk, abt, non, just, like, yes & 299,470 & 125.47 & 44,921 \\
\hline
\end{tabular}
}
\caption{Narrative archetype clusters with estimated conspiratorial message counts. Total estimated conspiratorial messages: 106,931.}
\label{tab:persona_clusters_full}
\vspace{-10pt}
\end{table*}

\begin{table}[h]
\centering
\sffamily
\resizebox{\columnwidth}{!}{
\begin{tabular}{p{0.4\textwidth}c}
\hline
\textbf{Model} & \textbf{Ground Truth Overlap (/9)} \\
\hline
\rowcollight \textbf{Hierarchical clustering with SiBeGNN embeddings} & 5/9 \\
Bertopic clusters with vanilla RoBERTa-large embeddings & 7/9 \\
\rowcollight Bertopic clustering with SiBeGNN embeddings & 6/9 \\
Hierarchical clusters with vanilla RoBERTa-large embeddings & 4/9 \\
\rowcollight HBScan clusters with vanilla RoBERTa-large embeddings & 2/9 \\
Gaussian mixture model with SiBeGNN embeddings & 5/9 \\
\rowcollight Spectral clustering with vanilla RoBERTa-large embeddings & 5/9 \\
KMeans with SiBeGNN embeddings & 5/9 \\
\hline
\end{tabular}
}
\caption{Comparison of narrative archetype coverage across clustering methods. Scores indicate the number of archetypes captured out of 9 possible ground truth narrative archetypes}
\label{tab:archetype_coverage_scores}
\vspace{-20pt}
\end{table}

\para{Manual verification of narrative archetypes: }
To evaluate the ability of different clustering algorithms to capture the diversity of narrative archetypes, we manually compared the clusters produced by each method against nine ground-truth archetypes defined in our study (\autoref{tab:archetype_coverage_scores}). Archetypes were defined by two experts who had published at least five articles on misinformation in Singapore and the broader region. Experts developed archetypes independently, and a third expert resolved disagreements. These archetypes represented distinct core narratives in the news and online environment during the study period. For instance, \textit{A1: Technocratic Control vs Personal Autonomy} reflects governance systems overriding individual choice, signaled by mentions of mandates, VDS compliance, surveillance, and bodily control. \textit{A2: Betrayal by Trusted Institutions} denotes perceived deception or incompetence by protective institutions, exemplified by government distrust or hospitals hiding data. \textit{A3: Unequal Citizenship \& Social Stratification} captures societal tiers with unequal rights, such as vaxxed versus unvaxxed populations and exclusion from jobs or venues. \textit{A4: Elite Collusion \& Loss of National Sovereignty} highlights decision-making dominated by elites or external powers, including global institutions or corporate capture. \textit{A5: Information Manipulation \& Manufactured Consent} focuses on media shaping opinion through censorship or framing, for example, news media distrust or propaganda. \textit{A6: Moral / Civilisational Decline} reflects the erosion of ethical boundaries, often involving religious framing, children, consent, or dehumanization. \textit{A7: Awakening, Resistance \& Minority Identity} describes minorities claiming awareness of hidden truths, signaled by terms like “sheeple,” “awakened,” or “truth-seekers.” \textit{A8: Everyday Life Under Systemic Stress} emphasizes the impact of large systems on daily life, such as disruptions to food, jobs, travel, healthcare access, or cost of living. Finally, \textit{A9: External Conflict as Moral Contrast} captures the use of foreign conflicts to interpret local events, for example, comparisons involving the US, China, or Russia.
The manual verification results show that \textbf{Hierarchical clustering with SiBeGNN embeddings} captured five of the nine archetypes, successfully identifying both everyday and conspiratorial discourse clusters and demonstrating strong separation of narrative personas. \textbf{Bertopic clusters with vanilla RoBERTa-large embeddings} captured seven archetypes, showing higher semantic coverage but lower interpretability due to some cluster overlap. \textbf{Bertopic clustering with SiBeGNN embeddings} identified six archetypes, balancing persona representation with semantic alignment. \textbf{Hierarchical clusters with vanilla RoBERTa-large embeddings} captured four archetypes, missing several conspiratorial narratives such as A2 and A6. \textbf{HBScan clusters with vanilla RoBERTa-large embeddings} captured only two archetypes, indicating poor granularity and failure to differentiate stylistic nuances. Other methods, including \textbf{Gaussian mixture model with SiBeGNN embeddings}, \textbf{Spectral clustering with vanilla RoBERTa-large embeddings}, and \textbf{KMeans with SiBeGNN embeddings}, captured five archetypes each, demonstrating moderate coverage but lower interpretability relative to hierarchical approaches.
These results indicate that \textbf{SiBeGNN-based hierarchical clustering} provides the best balance between coverage and interpretability. By leveraging disentangled, sign-aware embeddings, this approach effectively distinguishes both conspiratorial and everyday discourse modes, producing coherent and semantically meaningful narrative archetypes. In contrast, clustering methods that rely solely on vanilla embeddings or standard algorithms either fail to cover key archetypes or produce overlapping clusters, underscoring the importance of using belief-aware, persona-disentangled representations to capture the full spectrum of online discourse narratives.
\section{Discussion and Implications}

This study shows that conspiratorial discourse in Singapore-based Telegram groups is embedded within everyday communication rather than isolated spaces, challenging prevailing assumptions about online radicalization and highlighting the need to study such narratives within broader social discourse.

\para{Methodological implications:}
Our signed belief graph–based hierarchical clustering achieves the lowest cDBI (8.38) compared to standard methods (13.60–67.27), indicating more interpretable narrative archetypes~\cite{krasnov2019number}. The Sign-Disentanglement Loss effectively separates belief polarity from stylistic variation, extending beyond conspiracy detection to applications in stance detection, political discourse analysis, and other tasks requiring joint modeling of relational polarity and semantic content.

\para{Platform governance and generalizability:}
The seven archetypes exhibit distinct conspiracy prevalence patterns, indicating that moderation should be archetype-aware. For example, \textit{Medical Concerns} may benefit from targeted health interventions, while \textit{Contradictions in Authority} may require approaches addressing institutional trust deficits. Although the framework is platform- and region-agnostic, the identified archetypes reflect Singapore’s socio-political context, underscoring the need for cross-platform and cross-regional studies to distinguish universal from context-specific patterns.

\section{Conclusion}

This study shows that conspiratorial discourse in Singapore-based Telegram groups exists within everyday communication, spanning legal, health, media, and financial discussions, rather than in isolated echo chambers. Using a two-stage framework combining transformer-based classification and Signed Belief Graph Neural Networks (SiBeGNN), we identified seven narrative archetypes, revealing how conspiratorial content interweaves with ordinary social interaction.
The findings challenge assumptions about online radicalization: narrative archetypes from \textit{General Legal Topics} to \textit{Contradictions in Authority} illustrate that pragmatic, affective, and ideological communication coexist. Even skepticism-laden clusters appear within broader conversational contexts, highlighting the need for interventions that consider conspiratorial discourse’s embeddedness in everyday digital life.
Methodologically, the Sign-Disentanglement Loss separates belief polarity from narrative style, yielding superior clustering quality (cDBI = 8.38). This approach offers a replicable framework for studying belief-driven discourse across platforms, integrating signed network analysis with transformer embeddings to capture semantic content and relational structure, with applications in stance detection, and political discourse analysis.
\section{Limitations}

This study has several limitations presenting opportunities for future research. First, our analysis relies on textual embeddings from a specific transformer model; alternative architectures or multimodal data may reveal different patterns. Second, our normalization and weighting schemes, while theoretically justified, remain heuristic and could be optimized for specific objectives.
A critical limitation is our reliance on text-based features for archetype discovery. Narrative archetypes emerge from richer behavioral, temporal, and network-level dynamics beyond text alone. Notably, \textit{lurkers}, users who consume content without contributing, remain invisible to text-based analyses yet constitute substantial portions of communities and indirectly influence information ecosystems. Capturing such latent participation requires incorporating posting frequency, temporal activity patterns, engagement metrics, response latency, and structural network position (e.g., central versus peripheral actors). Future extensions incorporating multimodal or interaction-based signals could better model latent behavioral dimensions underlying archetype formation.
Finally, our dataset is restricted to Singapore-based Telegram groups and may not generalize to other regions or platforms with different cultural contexts, regulatory environments, or community norms. Future work could incorporate temporal dynamics of cluster evolution~\cite{shimgekar2025agentic, shimgekar2025nimblelabs}, integrate engagement metrics and social network structure, and conduct cross-platform comparative studies to identify universal versus context-specific patterns.

\section{Ethical considerations}

We implemented privacy safeguards including paraphrasing, anonymization, and analysis at the aggregate level. We acknowledge risks of surveillance misuse and emphasize that institutional skepticism is a core element of democratic discourse, distinct from false misinformation. Labeling health concerns as “conspiratorial” risks pathologizing legitimate grievances; our annotation protocols distinguish evidence-free claims from evidence-based criticism, though this boundary is context-dependent. We commit to public release of methods and code while restricting access to raw message data. The dataset may contain offensive or harmful language reflective of real-world discourse; we do not amplify such content and restrict access to the data in accordance with institutional ethical guidelines.

The labelling task required labeling each message as conspiratorial or non-conspiratorial based solely on message content, without attempting to infer author intent or user identity. As annotators were domain experts and collaborators on the project, no formal consent process or risk disclaimer was required; however, annotators were informed that the data may contain offensive or sensitive language typical of online political discourse. 



\bibliography{referencesCleaned}

\appendix
\clearpage
\appendix
\onecolumn  
\section{Appendix}
\setcounter{table}{0}
\setcounter{figure}{0}
\renewcommand{\thetable}{A\arabic{table}}
\renewcommand{\thefigure}{A\arabic{figure}}

\begin{table}[h]
\centering
\sffamily
\footnotesize
\resizebox{\columnwidth}{!}{%
\begin{tabular}{p{0.85\columnwidth} c}
\hline
\textbf{Text (masked for brevity)} & \textbf{Label} \\
\hline

\rowcollight Zionism -- A political movement that sucks the world dry.
The Synagogue of Satan consists of Jews and non-Jews.
(Biblical terminology hijacked strategically by Kazarian mafia/Mossad)
& 1 \\[6pt]

Long list of hawker centres temporarily closed for cleaning.
Some users speculate the word ``affected'' may imply closures targeting
unvaccinated patrons; others verify on NEA website that closures
are for cleaning only. Debate arises over tone and intention. [...]
& 0 \\[6pt]

\hline
\end{tabular}%
}
\caption{Masked chat excerpts with binary conspiratorial labels. Each example is labeled as conspiratorial (1) or not (0), with unimportant text masked using [...].}
\label{tab:masked_fullwidth}
\end{table}

\begin{table}[h]
\centering
\sffamily
\footnotesize
\resizebox{0.5\columnwidth}{!}{
\begin{tabular}{c c c c c c c}
\hline
\textbf{PCA} & \textbf{Th.} & \textbf{$k_{min}$} & \textbf{$k_{max}$} &
\textbf{Batch} & \textbf{State} & \textbf{Coher.} \\
\hline
\rowcollight 0.5 & 0.75 & 2 & 20 & 64 & 42 & 0.372 \\
0.5 & 0.75 & 3 & 20 & 64 & 42 & 0.381 \\
\rowcollight 0.5 & 0.75 & 4 & 20 & 64 & 42 & 0.365 \\
0.5 & 0.80 & 2 & 20 & 64 & 42 & 0.388 \\
\rowcollight 0.5 & 0.80 & 3 & 20 & 64 & 42 & 0.379 \\
0.5 & 0.80 & 4 & 20 & 64 & 42 & 0.384 \\
\rowcollight 0.6 & 0.75 & 2 & 20 & 64 & 42 & 0.376 \\
0.6 & 0.75 & 3 & 20 & 64 & 42 & 0.363 \\
\rowcollight 0.6 & 0.75 & 4 & 20 & 64 & 42 & 0.382 \\
\textbf{0.6} & \textbf{0.80} & \textbf{2} & \textbf{20} & \textbf{64} & \textbf{42} & \textbf{0.386} \\
\rowcollight 0.6 & 0.80 & 3 & 20 & 64 & 42 & 0.374 \\
0.6 & 0.80 & 4 & 20 & 64 & 42 & 0.386 \\
\rowcollight 0.7 & 0.75 & 2 & 20 & 64 & 42 & 0.368 \\
0.7 & 0.75 & 3 & 20 & 64 & 42 & 0.383 \\
\rowcollight 0.7 & 0.75 & 4 & 20 & 64 & 42 & 0.360 \\
0.7 & 0.80 & 2 & 20 & 64 & 42 & 0.371 \\
\rowcollight 0.7 & 0.80 & 3 & 20 & 64 & 42 & 0.378 \\
0.7 & 0.80 & 4 & 20 & 64 & 42 & 0.372 \\
\hline
\end{tabular}
}
\caption{Ablation results for hierarchical clustering parameters. Coherence values remain within 0.360--0.390, indicating stable behavior across parameter settings.}
\label{tab:hei_param}
\vspace{-20pt}
\end{table}

\begin{table}[h]
\centering
\sffamily
\small
\resizebox{\textwidth}{!}{
\begin{tabular}{lcccccc}
\hline
\textbf{Loss Variant} 
& \textbf{Avg. Coherence} 
& \textbf{Silhouette Score} 
& \textbf{Davies--Bouldin Index} 
& \textbf{cDBI} 
& $\boldsymbol{\Delta}$ \textbf{cDBI} \\
\hline

\rowcollight \textbf{Full SiBeGNN (all losses)} 
& 0.386 & -0.021 & 3.233 & \textbf{8.38} & --- \\

w/o $\mathcal{L}_{\text{orth}}$ (no orthogonality) 
& 0.341 & -0.028 & 4.521 & 13.26 & +58.2\% \\

\rowcollight w/o $\mathcal{L}_{\text{sign}}$ (no sign consistency) 
& 0.318 & -0.032 & 5.187 & 16.31 & +94.6\% \\

w/o $\mathcal{L}_{\text{belief}}$ (no belief alignment) 
& 0.297 & -0.037 & 5.894 & 19.85 & +136.9\% \\

\rowcollight w/o $\mathcal{L}_{\text{orth}}$ \& $\mathcal{L}_{\text{sign}}$ (both removed) 
& 0.289 & -0.041 & 6.523 & 22.57 & +169.3\% \\

Only $\mathcal{L}_{\text{recon}}$ (no task supervision) 
& 0.251 & -0.046 & 7.812 & 31.12 & +271.4\% \\

\rowcollight Vanilla RoBERTa (no GNN) 
& 0.234 & -0.015 & 8.246 & 35.24 & +320.5\% \\

\hline
\end{tabular}
}
\caption{Ablation study of sign-disentanglement loss components. Each row reports performance when specific loss terms are removed. $\Delta$ cDBI denotes percentage degradation relative to the full model. Lower cDBI values indicate better clustering quality.}
\label{tab:loss_ablation}
\end{table}

\end{document}